\documentclass[11pt]{article}
\usepackage{booktabs}
\usepackage{array}
\usepackage{rotating}
\usepackage{adjustbox}
\usepackage{multirow}
\usepackage[table]{xcolor}
\usepackage{caption}
\usepackage{hyperref}
\usepackage{amsmath}

\usepackage{hyperref}
\usepackage{url}
\hypersetup{hyperfootnotes=false}

\usepackage[final]{acl}

\usepackage{times}
\usepackage{latexsym}

\usepackage[T1]{fontenc}
\usepackage[utf8]{inputenc}

\usepackage{microtype}

\usepackage{inconsolata}

\usepackage{graphicx}

\title{DiZiNER: Disagreement-guided Instruction Refinement via Pilot Annotation Simulation for Zero-shot Named Entity Recognition}

\author{Siun Kim\thanks{This work was primarily conducted at the Biomedical Research Institute, Seoul National University Hospital.} \\
  Seltasquare \\
  Seoul, Korea \\
  \texttt{sukim@seltasquare.com} \\\And
  Hyung-Jin Yoon \\
  Seoul National University College of Medicine \\
  Seoul, Korea \\
  \texttt{hjyoon@snu.ac.kr} \\}

\begin{document}
\maketitle
\begin{abstract}
Large language models (LLMs) have advanced information extraction (IE) by enabling zero-shot and few-shot named entity recognition (NER), yet their generative outputs still show persistent and systematic errors. Despite progress through instruction fine-tuning, zero-shot NER still lags far behind supervised systems. These recurring errors mirror inconsistencies observed in early-stage human annotation processes that resolve disagreements through pilot annotation. Motivated by this analogy, we introduce DiZiNER (\textbf{Di}sagreement-guided \textbf{I}nstruction Refinement via Pilot Annotation Simulation for \textbf{Z}ero-shot \textbf{N}amed \textbf{E}ntity \textbf{R}ecognition), a framework that simulates the pilot annotation process, employing LLMs to act as both annotators and supervisors. Multiple heterogeneous LLMs annotate shared texts, and a supervisor model analyzes inter-model disagreements to refine task instructions. Across 18 benchmarks, DiZiNER achieves zero-shot SOTA results on 14 datasets, improving prior bests by +8.0 F1 and reducing the zero-shot to supervised gap by over +11 points. It also consistently outperforms its supervisor, GPT-5 mini, indicating that improvements stem from disagreement-guided instruction refinement rather than model capacity. Pairwise agreement between models shows a strong correlation with NER performance, further supporting this finding.\footnote{The code and prompts are available at \url{https://github.com/SiunKim/diziner-ner/}.}
\end{abstract}

\section{Introduction}
Information extraction (IE) converts unstructured text into structured data, with named entity recognition (NER) serving as the entry point that identifies and categorizes entity spans. Recent advances in large language models (LLMs) have greatly expanded the potential of IE (\citealp[]{lu2022unified, bogdanov2024nuner}), enabling in-context learning (ICL) strategies for NER such as few-shot (\citealp[]{chen2023learning, jiang2024p}) and zero-shot learning (\citealp[]{xie2023empirical, sainz2023gollie}). Despite this progress, state-of-the-art (SOTA) models still depend heavily on human-labeled data, with a wide performance gap remaining between supervised fine-tuning (SFT) and ICL (\citealp[]{xie2023empirical, naguib2024few}).

LLMs exhibit recurring NER error patterns, including difficulty following complex guidelines (\citealp[]{pang2023guideline, sainz2023gollie, qi2024adelie}), ambiguity in span boundary detection (\citealp[]{guo2024baner, ding2024rethinking}), and frequent confusion of entity types (\citealp[]{li2024re, kim2024verifiner}). Prior efforts have addressed these issues through instruction fine-tuning on diverse datasets (\citealp[]{wang2023gpt}), open NER frameworks (\citealp[]{sainz2023gollie}), and large-scale synthetic data generation (\citealp[]{zhou2023universalner}). Yet, supervised methods still outperform them by a considerable margin (Table \ref{tab:main_results}).

\begin{figure*}[t]
    \centering
    \includegraphics[width=\linewidth]{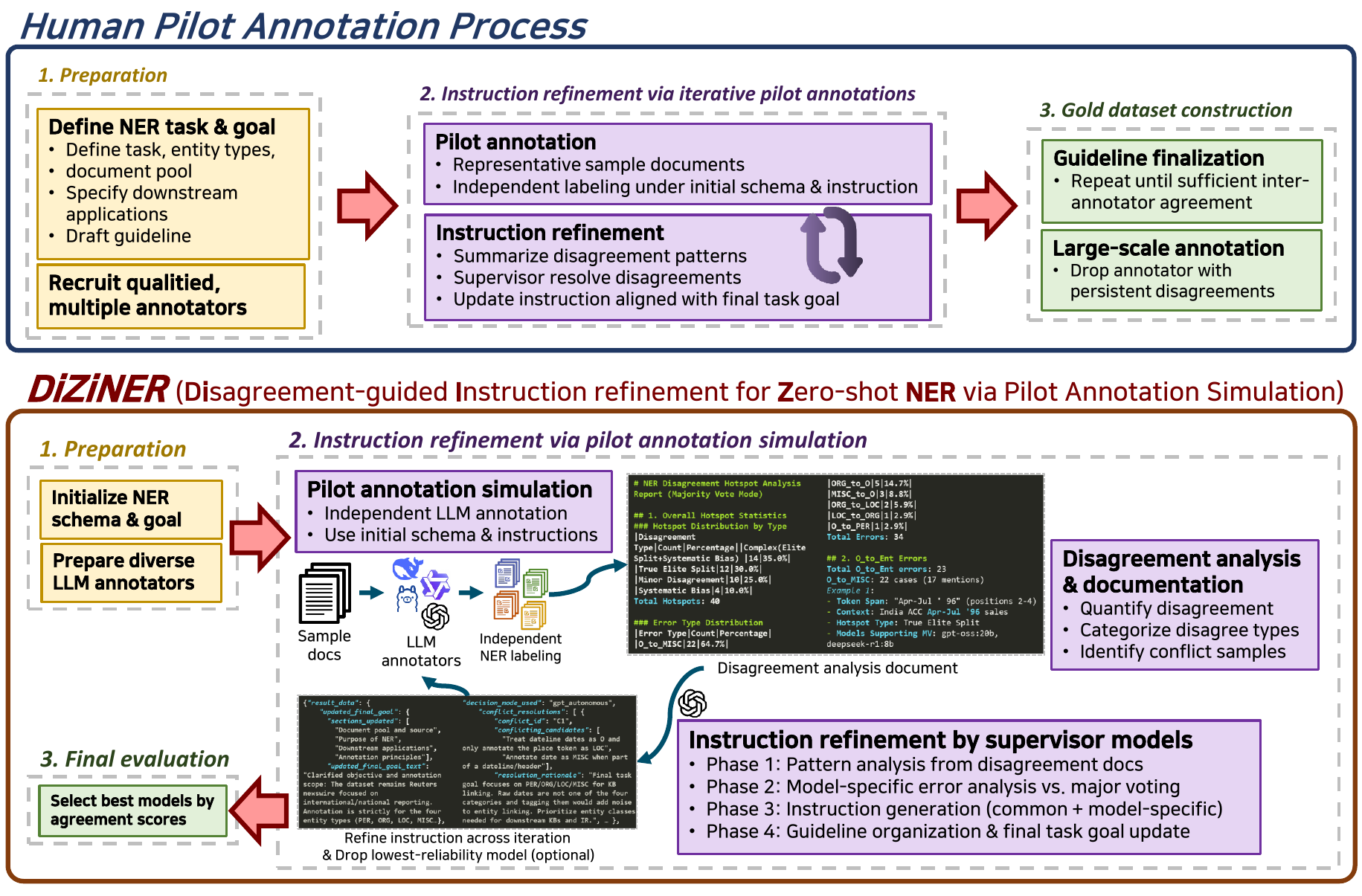}
    % \captionsetup{aboveskip=5pt}
    \caption{
        Overview of the DiZiNER framework. Multiple heterogeneous LLMs act as independent annotators.
        Disagreement profiles are constructed from their outputs, and a supervisor LLM iteratively refines
        the schema and annotator-specific instructions until convergence.
    }
    \label{fig:framework}
\end{figure*}

In this context, we note that these LLM errors parallel those observed during the early stages of human annotation (\citealp[]{tanabe2005genetag, bernier2024annotation}). Gold-standard datasets are typically built through \textit{pilot annotation}, an iterative process of resolving annotator disagreements and refining guidelines (\citealp[]{walker2006ace, weischedel2011ontonotes, finlayson2017overview}). Supervisors analyze disagreements, update ambiguous instructions, and align the annotations with downstream application needs (\citealp[]{fort2009towards}, Figure \ref{fig:framework}).

Building on this analogy, we propose DiZiNER (\textbf{Di}sagreement-guided \textbf{I}nstruction Refinement via Pilot Annotation Simulation for \textbf{Z}ero-shot \textbf{N}amed \textbf{E}ntity \textbf{R}ecognition), a framework that simulates pilot annotation using LLMs as both annotators and supervisors. Multiple heterogeneous open-source LLMs act as annotators labeling shared texts, and a supervisor LLM analyzes and categorizes inter-model disagreements to refine both common and model-specific instructions. This iterative cycle of annotation, disagreement analysis, and instruction refinement parallels the workflow of human pilot annotation, allowing LLMs to adapt to individual NER tasks without any parameter updates.

Across 18 NER benchmarks, DiZiNER achieves zero-shot SOTA results on 14 datasets, improving prior bests by +8.0 F1 on average and narrowing the gap between zero-shot and supervised performance from -32.0 to -20.9 points. Agreement metrics between LLM annotators consistently increase across iterations and show a strong correlation with NER performance. Notably, DiZiNER surpasses its GPT-5 mini supervisor, indicating that the observed improvements arise from disagreement-guided refinement rather than from the supervisor’s inherent capability.

\section{Related Works}
\paragraph{Instruction tuning for NER} Standard instruction fine-tuning often struggles to follow complex annotation guidelines and to produce structured outputs in IE tasks (\citealp[]{qi2024adelie}). InstructUIE and GoLLIE address these challenges by curating NER datasets for instruction fine-tuning, thereby improving zero-shot performance and guideline adherence (\citealp[]{wang2023instructuie, sainz2023gollie}). Open NER frameworks relax label constraints, allowing LLMs to better exploit their language understanding capabilities for NER (\citealp[]{etzioni2011open}). UniversalNER distills ChatGPT on synthetic data (\citealp[]{zhou2023universalner}), while GLiNER and NuNER adopt encoder-only architectures to reduce inference costs (\citealp[]{zaratiana2023gliner, bogdanov2024nuner}). Recent work has sought to unify heterogeneous corpora and to address span ambiguity through boundary-aware learning (\citealp[]{yang2024beyond, ding2024rethinking, guo2024baner}). Despite these advances, the performance gap with supervised systems remains large, and reliance on fine-tuning limits rapid adaptation to evolving LLMs.

\paragraph{Generative NER without instruction tuning}
In parallel, researchers have explored leveraging LLMs' inherent instruction-following capabilities to perform generative NER without requiring additional instruction fine-tuning. Early work constrained outputs via code-like schema representations (\citealp[]{li2023codeie, sainz2023gollie, guo2024retrieval, li2024knowcoder}) or reformulated tagging as token generation (\citealp[]{wang2023gpt}). Subsequent approaches introduced reasoning-based prompting such as self-consistency and self-verification methods to better convey complex annotation instructions (\citealp[]{xie2023empirical, kim2024verifiner, pang2023guideline}).

Building on the success of self-consistency and ICL, recent methods for generative NER adopt iterative self-improving strategies by generating pseudo-examples, filtering them, and providing them as in-context demonstrations (\citealp[]{xie2023self, tong2025evoprompt}). Our work follows this iterative, fine-tuning-free line of research yet distinctly utilizes inter-model disagreement as a signal for improving NER performance, paralleling how human annotators refine guidelines and reconcile judgments during gold-standard dataset construction.

\section{DiZiNER}
The DiZiNER framework operates through iterative pilot annotation cycles consisting of three stages: \textbf{(1) Independent Cross-Annotation}, where multiple LLM annotators independently perform NER tagging on the same set of documents; \textbf{(2) Disagreement Analysis}, which identifies \textit{hotspot} spans with high annotation disagreement, categorizes and summarizes disagreement patterns into structured reports; and \textbf{(3) Instruction Refinement}, where a supervisor model leverages the resulting structured reports to refine task instructions and reduces inter-model disagreement across iterations.

\subsection{Task Formulation}
LLM annotators form a heterogeneous pool $\mathcal{M}=\{M_k\}_{k=1}^{K}$ composed of independently developed models to minimize correlated errors. The label set is $\mathcal{L}=\{\ell_i\}_{i=1}^{n}$, and the NER schema is
\[
\Sigma=\big\{(\ell,\ d_\ell,\ \mathcal{P}_\ell,\ \mathcal{N}_\ell)\big\}_{\ell\in\mathcal{L}},
\]
where $d_\ell$ is a definition for entity type $\ell$, and $\mathcal{P}_\ell,\mathcal{N}_\ell$ are positive and negative examples. The schema $\Sigma$ remains fixed across iterations to maintain task consistency and prevent task drift.

At iteration $t$, annotator $M_k$ receives a task configuration
\[
\Theta_k^{(t)}=\big(\Sigma,\ C^{(t)},\ R_k^{(t)},\ G^{(t)}\big),
\]
where $C^{(t)}$ are common instructions, $R_k^{(t)}$ are model-specific instructions, and $G^{(t)}$ is the final task goal. Given an input sentence $x$, the annotator predicts
\[
y \sim P_{M_k}\!\big(y\,\big|\,x,\Theta_k^{(t)}\big),
\]
with labeled outputs $y=\{(e_j,\ \ell_{e_j})\}$, where $e_j$ is an entity span and $\ell_{e_j}\in\mathcal{L}$ denotes its label.

\subsection{Independent Cross-Annotation}
At each iteration, documents are grouped by lexical diversity, and a representative subset is randomly sampled across groups to form the iteration document set $\mathcal{D}^{(t)}$. All annotators in $\mathcal{M}$ independently label each sample in the set according to their task configuration $\Theta_k^{(t)}$. To enable token-level comparison across models, span-level annotations are converted into a BIO sequence representation. For input $x=(w_{1},\dots,w_{m})$, the tag set is defined as
\[
\mathcal{T}=\{\mathrm{B}\!-\!\ell,\ \mathrm{I}\!-\!\ell,\ \mathrm{O}\mid \ell\in\mathcal{L}\}.
\]
The conversion yields a BIO sequence
\[
\mathbf{z}_k(x)=(z_{k,1}(x),\dots,z_{k,m}(x)),\quad z_{k,i}(x)\in\mathcal{T},
\]
representing the token-level tagging output derived from the span-level annotation $y$ of annotator $M_k$.

\subsection{Disagreement Analysis}
This stage identifies \textit{hotspot} spans that exhibit strong inter-model disagreement. Token-level inconsistencies across annotators are quantified to mark high-disagreement regions.
\paragraph{Model Weights and Consensus}
Model weights are computed from pairwise strict span F1 scores between annotators, where for models $M_i$ and $M_j$,
\[
\mathrm{F1}_{ij}=\frac{2\,|\mathcal{S}_i\cap \mathcal{S}_j|}{|\mathcal{S}_i|+|\mathcal{S}_j|},
\]
where $\mathcal{S}_k$ denotes the set of predicted entity spans from model $M_k$.  
Each model’s weight, $w_k$, is computed as the average of its pairwise F1 scores with all others, normalized so that the weights sum to one. The \emph{elite set} is defined as the subset of annotators with the highest weights whose cumulative weight first reaches 0.5 when sorted in descending order. The computed model weights are also used as each annotator’s agreement score in subsequent analyses.

The consensus label for token $i$ in sentence $x$ is obtained via weighted majority voting,
\[
\widehat{\tau}(x,i)=\arg\max_{\tau\in\mathcal{T}}p_\tau(x,i),
\]
where \(p_\tau(x,i)=\sum_k w_k\,\mathbf{1}[z_{k,i}(x)=\tau]\) represents the weighted token-wise probability for tag $\tau$.

\paragraph{Hotspot Span Identification}
We compute three complementary token-level measures capturing distinct forms of annotation disagreement.  
(1) \textit{Label conflict} quantifies dispersion among BIO tags,
\[
D_{\mathrm{conf}}(x,i)=1-\sum_{\tau\in\mathcal{T}}p_\tau(x,i)^2.
\]
(2) \textit{Type confusion} reflects disagreement over entity types,
\[
D_{\mathrm{type}}(x,i) = 1 - \sum_{\ell \in \mathcal{L}} \left( \frac{p_{\mathrm{B}-\ell}(x,i) + p_{\mathrm{I}-\ell}(x,i)}{1 - p_{\mathrm{O}}(x,i)} \right)^2
\]
(3) \textit{Boundary uncertainty} measures inconsistency at entity boundaries,
\[
q_s(x,i) = \sum_{\ell \in \mathcal{L}} p_{\mathrm{B}-\ell}(x,i), \quad q_i(x,i) = \sum_{\ell \in \mathcal{L}} p_{\mathrm{I}-\ell}(x,i).
\]
\[
\begin{aligned}
U_{\mathrm{bnd}}(x,i) = \max \Big\{ & 4q_s(x,i)(1 - q_s(x,i)), \\
& 4q_i(x,i)(1 - q_i(x,i)) \Big\}.
\end{aligned}
\]

The final token-level disagreement score is defined as
\[
U_\star(x,i)=\max\{D_{\mathrm{conf}},\,D_{\mathrm{type}},\,U_{\mathrm{bnd}}\}.
\] Tokens are ranked by their $U_\star(x,i)$ scores, and the top 20\% are identified as high-disagreement regions. Neighboring tokens in this range are merged into hotspot spans, which are subsequently flagged for supervisor review and used for refining instructions.

\paragraph{Documentation}
Each iteration produces a summary report detailing hotspot statistics and model differences between elite and non-elite groups, outlining disagreement types and error categories (\textit{O$\rightarrow$Ent}, \textit{Ent$\rightarrow$O}, \textit{Ent$\rightarrow$Ent}, \textit{Span Error}). Representative examples with brief reasoning traces demonstrate characteristic disagreement patterns that inform targeted instruction refinement.

\begin{table*}[t]
\centering
\setlength{\tabcolsep}{6pt}
\begin{tabular}{@{}lcccccc@{}}
\toprule
\textbf{Methods} & \textbf{AI} & \textbf{Literature} & \textbf{Music} & \textbf{Politics} & \textbf{Science} & \textbf{Average} \\
\midrule
ChatGPT (\citealp[]{zhou2023universalner})   & 52.4 & 39.8 & 66.6 & 68.5 & 67.0 & 58.9 \\
GPT-4 (\citealp[]{yang2024beyond})           & 50.0 & 55.2 & 59.2 & 63.4 & 63.2 & 58.2 \\
InstructUIE (\citealp[]{wang2023instructuie}) & 49.0 & 47.2 & 53.2 & 48.1 & 49.2 & 49.3 \\
UniNER-7B (\citealp[]{zhou2023universalner})  & 53.6 & 59.3 & 67.0 & 60.9 & 61.1 & 60.4 \\
UniNER-13B (\citealp[]{zhou2023universalner}) & 54.2 & 60.9 & 64.5 & 61.4 & 63.5 & 60.9 \\
GLiNER (\citealp[]{zaratiana2023gliner})      & 57.2 & 64.4 & 69.6 & 72.6 & 62.6 & 65.3 \\
GoLLIE (\citealp[]{sainz2023gollie})          & 61.6 & 62.7 & 68.4 & 60.2 & 56.3 & 61.8 \\
KnowCoder-7B (\citealp[]{li2024knowcoder})    & 60.3 & 61.1 & 70.0 & 72.2 & 59.1 & 64.5 \\
IRRA (\citealp[]{xie2024retrieval})           & 57.5 & 59.3 & 69.4 & 74.0 & 68.3 & 65.7 \\
GNER (\citealp[]{ding2024rethinking})         & \underline{68.2} & 68.7 & \underline{81.2} & 75.1 & \underline{76.7} & 74.0 \\
B2NER (\citealp[]{yang2024beyond})            & 64.7 & \underline{71.6} & \textbf{82.4} & \underline{78.2} & \textbf{79.4} & \underline{75.3} \\
\midrule
GPT-5 mini (supervisor model)                 & 64.3 & 67.6 & 73.3 & 72.8 & 68.4 & 69.3 \\
\midrule
\textbf{DiZiNER}                 & \textbf{71.1} & \textbf{72.7} & 80.6 & \textbf{79.4} & 74.8 & \textbf{75.7} \\
\textbf{Avg. Gain from Iteration 0}           & \textbf{+2.7} & \textbf{+3.6} & \textbf{+11.1} & \textbf{+2.2} & \textbf{+4.5} & \textbf{+4.8} \\
\midrule
\textbf{\boldmath$\Delta_{\scriptsize \textit{DiZiNER - GPT-5 mini}}$} 
                                              & \textbf{+6.8} & \textbf{+5.1} & \textbf{+7.3} & \textbf{+6.6} & \textbf{+6.4} & \textbf{+6.4} \\
\textbf{\boldmath$\Delta_{\scriptsize \textit{DiZiNER - Best Prior Zero-shot}}$} 
                                              & \textbf{+2.9} & \textbf{+1.1} & \textbf{-1.8} & \textbf{+1.2} & \textbf{-4.6} & \textbf{-0.2} \\
\bottomrule
\end{tabular}
\caption{
Zero-shot NER performance on the CrossNER dataset. 
Best scores per domain are shown in \textbf{bold}, and second-best scores are \underline{underlined}. 
Avg. Gain from Iteration 0 denotes the mean improvement across eight annotator models, computed as the mean difference between each model’s Iteration-0 score and its best-performing iteration within the iterative document set.}
\label{tab:crossner_results}
\end{table*}

\subsection{Instruction Refinement}
The supervisor model iteratively refines task instructions based on disagreement documents and instructions from the previous iteration to improve annotator agreement. Each cycle proceeds through four phases:
\begin{enumerate}
    \item \textbf{Disagreement pattern analysis.}  
    Identify recurring disagreement patterns in hotspot summaries and infer their underlying causes. Extract generalizable correction principles rather than case-specific fixes (Table \ref{tab:phase1_prompt}).

    \item \textbf{Model-specific diagnosis.}  
    Examine residual errors for each non-elite model using the consensus output \(\widehat{\tau}(x,i)\), excluding patterns already addressed in Phase~1. Identify model-specific weaknesses and formulate targeted adjustments (Table \ref{tab:phase2_prompt}).

    \item \textbf{Guideline integration and conflict resolution.}  
    Integrate the refined instructions from the current supervision cycle with those from previous iterations, resolving any conflicts based on the final task goal, which aims to maximize performance in downstream applications (Table \ref{tab:phase3_prompt}).

    \item \textbf{Hierarchical organization.}  
    Reorganize refined instructions into a hierarchical structure where general rules precede specific or conditional cases. This restructuring enhances clarity and readability (Table \ref{tab:phase4_prompt}).
\end{enumerate}

A small set of tuning parameters was introduced to regulate the stability of iterative updates, and three parameter configurations were explored to ensure consistency across heterogeneous benchmarks (see Appendix~\ref{app:supervisor-parameters}). For supervised ablations, the same procedure is executed with gold-standard labels replacing the consensus outputs.

\subsection{Identification of Best Model Configuration}
DiZiNER selects the optimal iteration–model configuration, defined as the combination of a specific refinement iteration and an individual annotator model, in the absence of human-labeled data. We observe that pairwise annotator agreement, measured by strict span-level F1, is strongly correlated with NER performance (Figure~\ref{fig:agreement_correlation}), and thus use agreement statistics to guide configuration selection. Accordingly, all available iteration--model pairs are ranked by their mean agreement, and we report the average performance of the top three candidates on the test set.

\section{Experiments}
\subsection{Settings}
\paragraph{Datasets}  
We evaluate our framework on a total of 18 NER datasets spanning diverse domains, including the CrossNER suite (AI, Literature, Music, Politics, and Science; \citealp[]{liu2021crossner}), general-purpose corpora such as CoNLL2003 (\citealp[]{sang2003introduction}), ACE2005 (\citealp[]{walker2006ace}), OntoNotes (\citealp[]{pradhan2013towards}), and MultiNERD (\citealp[]{tedeschi2022multinerd}); biomedical corpora including AnatEM (\citealp[]{pyysalo2014anatomical}), BC2GM (\citealp[]{smith2008overview}), BC4CHEMD (\citealp[]{wang2019cross}), BC5CDR (\citealp[]{li2016biocreative}), and GENIA (\citealp[]{kim2003genia}); a STEM-oriented corpus FabNER (\citealp[]{kumar2022fabner}); and social or conversational datasets such as BroadTwitter (\citealp[]{derczynski2016broad}), MIT-Movie, and MIT-Restaurant (\citealp[]{liu2013asgard}). To simulate pilot annotation, we use the training splits to refine task instructions, while the final performance was evaluated on the corresponding test sets (Table \ref{tab:dataset_stats}).

\paragraph{Baselines}
We compare DiZiNER against representative baselines under both \textit{zero-shot} and \textit{supervised} settings.
The zero-shot setting excludes models that rely on task-specific fine-tuning or retrieval-based ICL, namely \textbf{ChatGPT} (\citealp{zhou2023universalner}), \textbf{GPT-4} (\citealp{yang2024beyond}), \textbf{InstructUIE} (\citealp{wang2023instructuie}), \textbf{UniNER-7B/13B} (\citealp{zhou2023universalner}), \textbf{GLiNER} (\citealp{zaratiana2023gliner}), \textbf{GoLLIE} (\citealp{sainz2023gollie}), \textbf{KnowCoder-7B} (\citealp{li2024knowcoder}), \textbf{GNER} (\citealp{ding2024rethinking}), \textbf{B2NER} (\citealp{yang2024beyond}), \textbf{IRRA} (\citealp{xie2024retrieval}), \textbf{EvoPrompt} (\citealp{tong2025evoprompt}), and \textbf{GPT-5 mini}.
For the supervised setting, we include SFT models trained on gold annotations, including \textbf{BERT-base} and \textbf{InstructUIE} (\citealp{wang2023instructuie}), \textbf{UniNER}, \textbf{GLiNER}, \textbf{KnowCoder-7B}, \textbf{GNER}, and \textbf{B2NER}.

\paragraph{Ensemble Baselines} To decouple the benefits of iterative refinement from potential ensemble effects, we compare DiZiNER against four consensus aggregation methods applied to the backbone models' initial outputs (Iteration 0). We include \textbf{Majority Voting (MV)}, \textbf{Dawid-Skene (DS)} \citep{dawid1979maximum}, \textbf{GLAD} \citep{whitehill2009whose}, and \textbf{MACE} \citep{hovy2013learning}. These baselines represent static "wisdom of the crowd" benchmarks, allowing us to isolate performance gains specifically attributable to our disagreement-guided instruction refinement process.

\paragraph{Backbones and Implementation}
DiZiNER employs a heterogeneous pool of eight open-source LLMs that were independently developed by different organizations, have distinct training architectures, datasets, and optimization pipelines, and are accessed via OpenRouter\footnote{\url{https://openrouter.ai}}:
\texttt{mistral-small3.2:24b},\hspace{0.1em}
\texttt{gpt-oss:20b},\hspace{0.1em}
\texttt{phi4:14b},\hspace{0.1em}
\texttt{qwen3:14b},\hspace{0.1em}
\texttt{gemma3:12b},\hspace{0.1em}
\texttt{deepseek-r1:8b},\hspace{0.1em}
\texttt{llama3.1:8b},\hspace{0.1em}
\texttt{nemotron-nano:8b}.
This diversity promotes independent judgment among annotators and minimizes correlated errors. The supervisor model was \texttt{GPT-5-mini-2025-08-07}, accessed via the OpenAI API between August~7 and September~30,~2025.

To ensure reproducibility and minimize variability from API-side updates, we utilized specific model snapshots (e.g., \texttt{llama-3.1-8b-instruct}) and a strictly deterministic decoding configuration: temperature 0.0, top-p 1.0, repetition penalty 1.0, and frequency/presence penalties 0.0, with a maximum output length of 8,000 tokens.

Each iteration processes a document set of 25 samples, with up to five refinement cycles. Three parameter configurations are explored to ensure consistent application across heterogeneous benchmarks (Table \ref{tab:tuning_configs}).

\paragraph{Metrics} We report the entity-level micro-F1 under the strict span setting as our evaluation metric, requiring both entity boundary and type to be correctly predicted.

\begin{table*}[t]
\centering
\begin{adjustbox}{width=\textwidth}
\begin{tabular}{@{}lcccccccccccccc@{}}
\toprule
\textbf{Methods} & \textbf{Movie} & \textbf{Rest.} & \textbf{B-Twit} & \textbf{ACE05} & \textbf{CoNLL} & \textbf{M-NERD} & \textbf{Onto} & \textbf{FabNER} & \textbf{Anat} & \textbf{bc2} & \textbf{bc4} & \textbf{bc5} & \textbf{GENIA} & \textbf{Avg} \\
\midrule
\multicolumn{15}{@{}l}{\textbf{\textit{Zero-shot}}} \\ 
\hspace{0.5em}ChatGPT (\citealp[]{zhou2023universalner}) & 5.3 & 32.8 & 61.8 & 26.6 & 52.5 & 58.1 & 29.7 & 15.3 & 30.7 & 40.2 & 35.5 & 52.4 & 41.6 & 37.1 \\
\hspace{0.5em}GoLLIE (\citealp[]{sainz2023gollie}) & 63.0 & 52.7 & 51.4 & -- & -- & \underline{77.5} & -- & \underline{26.3} & -- & -- & -- & -- & -- & -- \\
\hspace{0.5em}UniNER-7B (\citealp[]{zhou2023universalner}) & 42.4 & 31.7 & \underline{67.9} & 36.9 & 72.2 & 59.3 & 27.8 & 24.8 & 25.1 & 46.2 & \underline{47.9} & \underline{68.0} & 54.1 & 46.5 \\
\hspace{0.5em}GLiNER (\citealp[]{zaratiana2023gliner}) & 57.2 & 42.9 & 61.2 & 27.3 & 64.6 & 59.7 & \underline{32.2} & 23.6 & \underline{33.3} & \underline{47.9} & 43.1 & 66.4 & \underline{55.5} & \underline{47.3} \\
\hspace{0.5em}EvoPrompt (\citealp[]{tong2025evoprompt}) & \underline{70.9} & \textbf{69.3} & -- & \textbf{51.2} & \underline{81.3} & -- & -- & -- & -- & -- & -- & -- & -- & -- \\
\midrule
GPT-5 mini (supervisor model) & 73.3 & 58.5 & 59.2 & 54.0 & 81.8 & 74.0 & 63.8 & 29.5 & 59.2 & 73.0 & 63.7 & 62.8 & 56.5 & 62.3 \\ 
\midrule
\textbf{DiZiNER} & \textbf{76.2} & \underline{67.3} & \textbf{76.9} & \underline{45.0} & \textbf{86.9} & \textbf{80.6} & \textbf{62.5} & \textbf{29.5} & \textbf{59.1} & \textbf{71.0} & \textbf{79.5} & \textbf{78.9} & \textbf{60.1} & \textbf{68.4} \\
\textbf{Avg. Gain from Iteration 0} & \textbf{+5.6} & \textbf{+22.9} & \textbf{+20.1} & \textbf{+2.1} & \textbf{+28.3} & \textbf{+2.6} & \textbf{+24.8} & \textbf{+0.9} & \textbf{+25.3} & \textbf{+12.7} & \textbf{+26.9} & \textbf{+16.4} & \textbf{+4.5} & \textbf{+14.9} \\
\textbf{\boldmath$\Delta_{\textit{DiZiNER - GPT-5 mini}}$} & \textbf{+2.9} & \textbf{+8.8} & \textbf{+17.7} & \textbf{-9.0} & \textbf{+5.1} & \textbf{+6.6} & \textbf{-1.3} & \textbf{+0.0} & \textbf{-0.1} & \textbf{-2.0} & \textbf{+15.8} & \textbf{+16.1} & \textbf{+3.6} & \textbf{+5.0} \\
\midrule
\multicolumn{15}{@{}l}{\textbf{\textit{Supervised}}} \\ 
\hspace{0.5em}BERT-base (\citealp[]{wang2023instructuie}) & 88.8 & 81.0 & 58.6 & \textbf{87.3} & 92.4 & 91.3 & \underline{91.1} & 64.2 & 85.8 & 80.9 & 86.7 & 85.3 & 73.3 & 82.1 \\
\hspace{0.5em}InstructUIE (\citealp[]{wang2023instructuie}) & 89.6 & 82.6 & 80.3 & 79.9 & 91.5 & 90.3 & 88.6 & 78.4 & 88.5 & 80.7 & 87.6 & 89.0 & 75.7 & 84.8 \\
\hspace{0.5em}UniNER-7B (\citealp[]{zhou2023universalner}) & 90.2 & 82.3 & 81.2 & \underline{86.7} & 93.3 & 93.7 & 89.9 & 81.9 & 88.5 & 82.4 & \underline{89.2} & \underline{89.3} & \underline{77.5} & \textbf{86.6} \\
\hspace{0.5em}GLiNER (\citealp[]{zaratiana2023gliner}) & 87.9 & 83.6 & \textbf{82.7} & 82.8 & 92.6 & \textbf{93.8} & 89.0 & 77.8 & 88.9 & \underline{83.7} & 87.9 & 88.7 & \textbf{78.9} & \underline{86.0} \\
\hspace{0.5em}KnowCoder-7B (\citealp[]{li2024knowcoder}) & \underline{90.6} & 81.3 & 78.3 & 86.1 & \textbf{95.1} & 93.1 & 88.2 & \underline{82.9} & 86.4 & \underline{82.0} & -- & \underline{89.3} & 76.7 & -- \\
\hspace{0.5em}GNER (\citealp[]{ding2024rethinking}) & 90.2 & \textbf{83.8} & 81.3 & -- & \underline{93.6} & \textbf{94.4} & \textbf{91.8} & \textbf{85.4} & \textbf{90.3} & \textbf{84.3} & \textbf{90.0} & \textbf{90.3} & -- & -- \\
\hspace{0.5em}B2NER (\citealp[]{yang2024beyond}) & \textbf{90.8} & \underline{83.7} & \underline{82.2} & 83.0 & 92.6 & \underline{94.0} & 84.3 & 78.8 & \underline{89.2} & 82.0 & 89.0 & 88.5 & 76.4 & 85.7 \\
\midrule
\textbf{\boldmath$\Delta_{\textit{Best prior ZS - Best prior Sup.}}$} & \textbf{-19.9} & \textbf{-14.5} & \textbf{-14.8} & \textbf{-36.1} & \textbf{-13.8} & \textbf{-16.9} & \textbf{-59.6} & \textbf{-59.1} & \textbf{-57.0} & \textbf{-36.4} & \textbf{-42.1} & \textbf{-22.3} & \textbf{-23.4} & \textbf{-32.0} \\ 
\textbf{\boldmath$\Delta_{\textit{DiZiNER - Best prior ZS}}$} & \textbf{+5.3} & \textbf{-2.0} & \textbf{+9.0} & \textbf{-6.2} & \textbf{+5.6} & \textbf{+3.1} & \textbf{+30.3} & \textbf{+3.2} & \textbf{+25.8} & \textbf{+23.1} & \textbf{+31.6} & \textbf{+10.9} & \textbf{+4.6} & \textbf{+11.1} \\ 
\textbf{\boldmath$\Delta_{\textit{DiZiNER - Best prior Sup.}}$} & \textbf{-14.6} & \textbf{-16.5} & \textbf{-5.8} & \textbf{-42.3} & \textbf{-8.2} & \textbf{-13.8} & \textbf{-29.3} & \textbf{-55.9} & \textbf{-31.2} & \textbf{-13.3} & \textbf{-10.5} & \textbf{-11.4} & \textbf{-18.8} & \textbf{-20.9} \\ 
\bottomrule
\end{tabular}
\end{adjustbox}
\caption{Overall NER results across 13 benchmarks. \emph{ZS} denotes our zero-shot pipeline without any gold labels, and all DiZiNER results are zero-shot. Within each setting (zero-shot and supervised), the best and second-best scores are highlighted in \textbf{bold} and \underline{underlined}, respectively. GPT-5 mini results are excluded from this comparison. Avg. Gain from Iteration 0 denotes the average improvement averaged across eight backbone models, computed as the mean difference between each model’s Iteration-0 score and its best-performing iteration within the iterative document set. Overall performance is averaged only for models with complete results across all benchmarks. Abbreviations: Movie = MIT-Movie, Rest. = MIT-Restaurant, B-Twit = BroadTwitter, ACE05 = ACE2005, CoNLL = CoNLL2003, M-NERD = MultiNERD, Onto = OntoNotes, Anat = AnatEM, bc2 = BC2GM, bc4 = BC4CHEMD, bc5 = BC5CDR.}
\label{tab:main_results}
\end{table*}

\subsection{Main Results}
Without any instruction fine-tuning, DiZiNER establishes new zero-shot SOTA results on 14 out of 18 benchmarks (Tables~\ref{tab:crossner_results} and~\ref{tab:main_results}). 
On CrossNER, DiZiNER achieves SOTA performance in three of the five domains, excluding Music and Science, with an average F1 of 75.7, outperforming B2NER~\citep[]{yang2024beyond} by +0.4 F1 points. In addition, compared with its GPT-5 mini supervisor, DiZiNER yields an average improvement of +6.4 F1 (Table~\ref{tab:crossner_results}).

Across benchmarks with available supervised results, DiZiNER improves the average zero-shot performance by +11.1 F1 points over the best prior zero-shot and narrows the gap between zero-shot and supervised performance from -32.0 to -20.9 F1 (Table \ref{tab:main_results}). DiZiNER surpasses its GPT-5 mini supervisor by an average of +5.0 F1 points, demonstrating that the observed improvements arise from disagreement-guided refinement rather than the supervisor’s intrinsic capability.

DiZiNER averages 69.6 on CrossNER AI and Literature, outperforming the four static ensemble aggregators (Table \ref{tab:ensemble_comparison_final}). While these ensembles, including Majority Voting (66.6), already surpass prior zero-shot SOTA, DiZiNER consistently exceeds the strongest method, MACE (67.0). This confirms that iterative refinement is essential for driving performance gains beyond the reach of static consensus alone.

In practice, NER performance on the iteration document sets consistently improved across iterations. When averaged over the eight annotator models, performance increased from Iteration 0 to each model’s best-performing iteration by as much as +25 F1 on several benchmarks, with an overall average gain of +14.9 F1 (Table \ref{tab:main_results}) and +4.8 F1 on CrossNER (Table \ref{tab:crossner_results}).

Individual LLM annotators' performance generally improves through refinement, a trend closely tracked by inter-model agreement. While performance typically peaks at 2.7 iterations on average, trajectories vary significantly across benchmarks (Figure \ref{fig:performance_evolution_full}). Early peaks (e.g., MIT-Movie) can decline due to overcorrection from the fixed 20\% threshold, while complex or high-density tasks like OntoNotes 5.0 and Broad Twitter exhibit more gradual or volatile patterns.

Notably, despite these diverse trajectories, inter-model agreement remains a consistently reliable proxy for NER performance. This reliability is substantiated by strong F1-agreement correlations, reaching $\rho=0.922$ for CrossNER-Politics and $0.886$ for OntoNotes 5.0 (Figure \ref{fig:agreement_correlation}). This relationship validates model consensus as a robust, label-free indicator of task quality, supporting the effectiveness of the DiZiNER framework.

Sensitivity analysis across five seeds confirms the framework’s robustness to stochasticity in both token sampling and refinement pathways, yielding low standard deviations of 0.8\% and 2.1\% for CrossNER-AI and Literature, respectively. Furthermore, while precise instruction crafting is beneficial for maximizing performance, evaluations across five distinct initial instructions yield mean F1 scores of 67.2\% [65.7\%, 69.9\%] for CrossNER-AI and 70.4\% [69.1\%, 72.7\%] for Literature, confirming the framework's stability against variations in the initial task instructions.

The average cost per iteration was \$1.90 for inference and \$0.77 for supervision, resulting in a total of \$2.67 per iteration. Considering that an average of five iterations were conducted for each benchmark and three configuration settings were explored, the total cost amounts to \$40.1 per benchmark.

\subsection{Ablation Study}
\paragraph{Annotator Diversity and Scaling}
Diverse ensembles of smaller models ($\le$ 24B) consistently outperform single-family pools by 1.7--3.7 F1 points despite the latter’s larger scale (Table~\ref{tab:abl_llm_families}). Scaling from 4 to 8 annotators improves average F1 from 73.1 to 75.5, yet performance declines beyond 12 models (73.9) due to increased consensus noise (Table~\ref{tab:performance_model_comparison}). Consequently, we recommend employing a heterogeneous pool of 8--12 annotator models from distinct lineages to optimally balance signal diversity and consensus stability.

\paragraph{Supervisor model capacity.}
Evaluation across diverse high-capacity supervisors shows consistent improvements over the GPT-5 mini baseline, though a performance gap remains compared with the prior zero-shot SOTA (Table \ref{tab:diverse_supervisors}). These findings suggest that while the disagreement-guided refinement is effective across various models, the supervisor’s capability remains a relevant factor in determining the final performance levels achieved by the framework.

\paragraph{Final task goal}
Skipping the final task goal consistently degraded performance, leading to a significant average F1 drop from 77.6 to 71.9 across CrossNER and CoNLL2003 (Table~\ref{tab:skipping_final_goal}). In our framework, this component was designed to serve as a global criterion that guides instruction refinement toward the overall task objective. We speculate that when it is omitted, the refined instructions may remain locally consistent yet diverge from the benchmark’s intended direction, leading to lower F1 scores across domains.

\paragraph{Removing the least consistent annotator}
The effect of excluding the most disagreement-prone annotator varied across benchmarks, with no consistent trend (Table~\ref{tab:abl_dropping_worst}). Removing the least consistent annotator sometimes improved results but also risked destabilizing disagreement statistics by reducing diversity. Given these trade-offs, we treat this step as a tunable option rather than a fixed rule.

\paragraph{Iteration set size}
Optimal performance was achieved with 15--25 samples (Table~\ref{tab:abl_group_size}), while larger sets degraded results by expanding hotspot regions and obscuring distinct error patterns. Further scaling of the iteration document set size appears unnecessary under the current framework.

\paragraph{DiZiNER with gold-standard data}
Incorporating gold supervision provided minimal benefits in our framework (Table~\ref{tab:results_supervised}). Average performance increased by 0.3 F1, with consistent gains observed on ACE05 (+10.5) and OntoNotes (+5.6), where human annotations helped resolve errors arising from missing context and pronominal references. Replacing disagreement signals with gold labels shifted the objective from cross-model consensus to fixed-target fitting, thereby reducing diversity and weakening iterative refinement. Overall, the disagreement-guided setup without supervision achieved greater stability and stronger performance.

\subsection{Instruction Refinement and NER Quality Analysis}
Instruction analysis (Appendix \ref{app:instruction_analysis}) reveals that span boundary, entityhood, and type disambiguation constitute approximately 60\% of all refined instructions (Table \ref{tab:instruction_category_stats}). This concentration is consistent with prior observations of human annotator disagreements during pilot annotation, confirming that the simulated refinement effectively targets established bottlenecks. High-performing configurations further distinguish themselves by emphasizing global strategy (+2.8\%) and entityhood (+4.7\%) to better align with task objectives.

These refinements are qualitatively evident in DiZiNER’s ability to address persistent errors by synthesizing valid rules from document-level signals (Table \ref{tab:qualitative_ner_results}). For instance, by leveraging contextual cues, DiZiNER correctly classifies "Cambridge" as an organization within league tables and recovers previously missed publication names like "Nature," ensuring domain-wide consistency through instruction-based signal discovery.

\section{Conclusion}
We introduce DiZiNER, a zero-shot NER framework that simulates human pilot annotation through disagreement-guided instruction refinement without any parameter updates. By employing multiple heterogeneous LLMs as annotators and a supervisor model for disagreement-driven refinement, DiZiNER reduces boundary ambiguity and type confusion. Across 18 benchmarks, it achieves zero-shot SOTA results on 14 datasets, improves over the previous best zero-shot systems by +11.1 F1 points on average, outperforms its GPT-5 mini supervisor, and narrows the zero-shot-to-supervised gap from -32.0 to -20.9 F1 points. Ablation studies show that aligning refinement with the final task objective is essential for resolving conflicting instructions and that annotator diversity is critical for effective updates. The strong correlation between agreement metrics and gold-standard F1 indicates that disagreement-guided refinement is the primary driver of gains, suggesting that small open-source models can often surpass advanced proprietary baselines in a fully zero-shot setting without instruction fine-tuning.

\section*{Limitations}
Our framework exhibits varying gains across benchmarks. This variability likely stems from stochasticity and sampling differences that can alter the trajectory of iterative refinement. Because DiZiNER represents each dataset through its document pool and NER schema without accessing gold-labeled samples, refined instructions may gradually drift from dataset-specific annotation conventions. A hybrid approach that combines disagreement-guided refinement with a small number of supervised examples could anchor the process to the intended labeling criteria while preserving the efficiency and generality of zero-shot learning.

We also keep the NER schema fixed across iterations to maintain comparability and evaluation consistency with benchmark datasets. This design departs from realistic pilot annotation workflows, where entity types are often added, merged, or removed to resolve ambiguities and better capture domain semantics. Extending DiZiNER to real-world corpus construction will therefore require a schema-refinement component that can propose, test, and validate type updates while ensuring backward compatibility with earlier iterations and maintaining evaluation continuity.

\section*{Acknowledgments}

This work was supported by the Institute of Information \& Communications Technology Planning \& Evaluation(IITP)-Innovative Human Resource Development for Local Intellectualization program grant funded by the Korea government(MSIT)(IITP-2026-RS-2024-00441407)

During the preparation of this paper, we utilized ChatGPT and Gemini for grammatical proofreading, and Claude and GitHub Copilot to assist with code generation. All AI-generated code was rigorously reviewed, tested, and publicly released to ensure reproducibility. The authors take full responsibility for all content and results presented in this work.

\bibliography{custom}
\appendix
\section{Tuning Parameters for Instruction Refinement}
\label{app:supervisor-parameters}

To stabilize instruction refinement and prevent excessive corrections during iterative updates, the supervisor employs a series of tuning parameters. Each parameter controls a distinct aspect of the refinement process, ensuring balanced evolution of the instruction set across different phases.

\paragraph{Parameter Definitions}
\begin{itemize}
    \item \textbf{\texttt{max\_common\_instructions}}  
    Specifies the maximum number of newly generated shared principles per iteration. This prevents uncontrolled guideline expansion and is primarily active in \textbf{Phase~1} and \textbf{Phase~3}.

    \item \textbf{\texttt{max\_patterns}}  
    Determines the number of disagreement patterns considered in each cycle. By focusing only on the most recurrent inconsistencies, it guides efficient refinement during \textbf{Phase~1}.

    \item \textbf{\texttt{max\_model\_specific\_instructions}}  
    Sets an upper bound on model-specific adjustments per annotator model. This maintains an appropriate balance between general and specialized rules in \textbf{Phase~2} and \textbf{Phase~3}.

    \item \textbf{\texttt{limit\_instruction\_changes}}  
    Enables a controlled edit mode that constrains the degree of revision between refinement cycles, applied during \textbf{Phase~4}.

    \item \textbf{\texttt{max\_change\_ratio}}  
    When controlled editing is active, this parameter limits the proportion of textual modifications to preserve continuity and prevent semantic drift, also enforced in \textbf{Phase~4}.
\end{itemize}

\paragraph{Representative Configurations}
Three parameter configurations were adopted across benchmarks to investigate varying levels of refinement adaptiveness (Table \ref{tab:tuning_configs}). All settings used a group size of 25 samples per iteration and a maximum of four refinement cycles.

\begin{table*}[t]
\centering
\renewcommand{\arraystretch}{1.1}
\begin{adjustbox}{max width=\textwidth}
\begin{tabular}{@{}lccccccc@{}}
\toprule
\textbf{Configuration} & \textbf{max\_common} & \textbf{max\_patterns} & \textbf{max\_model\_spec.} & \textbf{limit\_changes} & \textbf{max\_ratio} & \textbf{max\_iter.} \\ 
\midrule
Stable & 3 & 5 & 2 & True & 0.10 & 5 \\
Relaxed & 5 & 8 & 3 & False & 0.20 & 5 \\
Aggressive & 10 & 20 & 10 & False & 0.50 & 5 \\
\bottomrule
\end{tabular}
\end{adjustbox}
\caption{Tuning configurations for instruction refinement experiments.}
\label{tab:tuning_configs}
\end{table*}

\section{Dataset Statistics}
\label{app:dataset_stats}
Table~\ref{tab:dataset_stats} summarizes the datasets used across experiments, encompassing 18 NER benchmarks from general, biomedical, STEM, and social domains to ensure broad domain coverage and diversity of entity types.

\begin{table*}[t]
\centering
\small
\resizebox{\textwidth}{!}{
\begin{tabular}{>{\raggedright\arraybackslash}m{3.0cm} l c c c c c c}
\toprule
\textbf{Domain} & \textbf{Dataset} & \textbf{\# train} & \textbf{\# dev} & \textbf{\# test} & \textbf{\# types} & \textbf{Avg. tokens} & \textbf{Avg. entities} \\
\midrule
\multirow{5}{=}{\textbf{Cross-domain (CrossNER)}} 
& AI (\citealp[]{liu2021crossner}) & 100 & 350 & 431 & 14 & 52 & 5.3 \\
& Literature (\citealp[]{liu2021crossner}) & 100 & 400 & 416 & 12 & 54 & 5.4 \\
& Music (\citealp[]{liu2021crossner}) & 100 & 380 & 465 & 13 & 57 & 6.5 \\
& Politics (\citealp[]{liu2021crossner}) & 199 & 540 & 650 & 9 & 61 & 6.5 \\
& Science (\citealp[]{liu2021crossner}) & 200 & 450 & 543 & 17 & 54 & 5.4 \\
\midrule
\multirow{3}{=}{\textbf{Social Media / Dialogue}}
& MIT-Movie (\citealp[]{liu2013asgard}) & 9775 & 2442 & 2443 & 12 & 10 & 2.2 \\
& MIT-Restaurant (\citealp[]{liu2013asgard}) & 7660 & 1520 & 1521 & 8 & 9 & 2.0 \\
& BroadTwitter (\citealp[]{derczynski2016broad}) & 5334 & 2001 & 2000 & 3 & 28 & 0.5 \\
\midrule
\multirow{4}{=}{\textbf{General}}
& ACE2005 (\citealp[]{walker2006ace}) & 7299 & 971 & 1060 & 7 & 21 & 2.8 \\
& CoNLL2003 (\citealp[]{sang2003introduction}) & 14041 & 3250 & 3453 & 4 & 25 & 2.8 \\
& MultiNERD (\citealp[]{tedeschi2022multinerd}) & 134144 & 10000 & 10000 & 16 & 28 & 1.6 \\
& OntoNotes (\citealp[]{pradhan2013towards}) & 59924 & 8528 & 8262 & 18 & 18 & 0.9 \\
\midrule
\multirow{1}{=}{\textbf{STEM}}
& FabNER (\citealp[]{kumar2022fabner}) & 9435 & 2182 & 2064 & 12 & 36 & 5.1 \\
\midrule
\multirow{4}{=}{\textbf{Biomedical}}
& AnatEM (\citealp[]{pyysalo2014anatomical}) & 5861 & 2118 & 3830 & 1 & 37 & 0.7 \\
& BC2GM (\citealp[]{smith2008overview}) & 12500 & 2500 & 5000 & 1 & 36 & 0.4 \\
& BC4CHEMD (\citealp[]{wang2019cross}) & 30682 & 30639 & 26364 & 1 & 45 & 0.9 \\
& BC5CDR (\citealp[]{li2016biocreative}) & 4560 & 4581 & 4797 & 2 & 41 & 2.2 \\
& GENIA (\citealp[]{kim2003genia}) & 15022 & 1669 & 1855 & 5 & 46 & 3.5 \\
\bottomrule
\end{tabular}
}
\caption{Statistics of datasets used in our experiments. We evaluate across 18 NER datasets covering general, biomedical, STEM, and social domains.}
\label{tab:dataset_stats}
\end{table*}

\section{Prompts}
Tables \ref{tab:phase1_prompt}-\ref{tab:phase4_prompt} summarize the four supervisory prompts used for instruction refinement: disagreement analysis, model-specific error review, instruction generation, and hierarchical organization. Each phase builds on the previous to ensure consistent, interpretable NER annotation. Full prompt templates and JSON schema are available on the project’s GitHub repository.

\begin{table*}[t!]
\centering
\rowcolors{1}{lightgray}{white}
\begin{tabular}{p{0.96\textwidth}}
\toprule
\textbf{Prompt Text} \\
\midrule
You are a strict, methodical NER annotation supervisor. Your task in this phase is to analyze disagreement patterns using majority-vote as a reference point (NOT as ground truth), identify high-yield error patterns, and classify them systematically. Focus on extracting actionable patterns that can inform instruction creation without generating instructions yet.

\textbf{Current NER Scheme:} \texttt{\{current\_ner\_schema\}}.

\textbf{Final Task Goal:} \texttt{\{final\_task\_goal}\}.

\textbf{Disagreement Analysis:} \texttt{\{NER\_disagreement\_summaries}\}.

\textbf{Task:} 
1. Analyze disagreement patterns using MV as reference point, acknowledging that MV is not ground truth but a useful consensus measure.
2. Identify and quantify disagreement patterns, clustering them into maximum 8 high-impact categories.
3. For each pattern, determine root causes and assess whether existing instructions already address them.
4. Extract possible annotation approaches for each conflicting case, providing the rationale behind each approach.
5. Identify aspects of the final task goal that need clarification to resolve ambiguous annotation choices.
6. Do NOT generate instructions in this phase - focus on pattern analysis and candidate instruction principles.

\textbf{Output Format (JSON)}:
\begin{verbatim}
{
  "disagreement_analysis_summary": {
    "major_disagreement_sources": [
      "Source 1", ...
    ],
    "mv_reference_reliability": "Assessment of MV as reference point",
    "elite_vs_non_elite_patterns": 
    "Comparison between elite and non-elite model behaviors"
  },
  "identified_patterns": [
    {
      "pattern_id": "P1",
      "pattern_name": "Descriptive pattern name",
      "frequency": "high|medium|low",
      "disagreement_subtypes": [
        "Subtype A", ...
      ],
      "root_cause_analysis": 
      "Fundamental principle-level explanation of disagreement source",
      "affected_entity_types": [
        "PER", ...
      ],
      "annotation_approaches": [
        {
          "approach": "Annotation approach A",
          "rationale": "Why this approach makes sense",
          "supporting_models": [
            "model1", ...
          ]
        }, ...
      ], ...
  ]
}\end{verbatim}\\
\bottomrule
\end{tabular}
\caption{Prompt for instruction refinement in phase 1 (Disagreement Pattern Analysis).}
\label{tab:phase1_prompt}
\end{table*}

\begin{table*}[t!]
\centering
\rowcolors{1}{lightgray}{white}
\begin{tabular}{p{0.96\textwidth}}
\toprule
\textbf{Prompt Text}\\
\midrule
You are a strict, methodical NER annotation supervisor. In this phase, you analyze error patterns specific to \texttt{{model\_name}} based on the configuration, focusing on systematic deviations not covered by common disagreement patterns identified in Phase 1. IMPORTANT: You are analyzing only one model at a time, not multiple models.

\textbf{Inputs} 

- Phase 1 results: \texttt{{phase1\_results}} 

- Single model detailed disagreement data: \texttt{{model\_disagreement\_data}}

- Elite model identification results: \texttt{{elite\_models}} 

- Single model bias analysis: \texttt{{model\_bias\_analysis}} 

- Original NER scheme: \texttt{{current\_ner\_schema}} 

- Final task goal: \texttt{{final\_task\_goal}} 

- Existing model-specific instructions: \texttt{{existing\_model\_instructions}} 

\textbf{Runtime Identifiers:} \texttt{{model\_name}}, \texttt{{model\_type}}\

\textbf{Task Focus} 

- Identify model-unique patterns not covered by Phase 1 

- Classify into \texttt{systematic\_bias | confusion\_pattern | under\_tagging | over\_tagging | boundary\_errors} 

- Assess existing model-specific instructions and their effectiveness 

- Prepare instruction \emph{needs} (do not generate instructions yet)

\textbf{Output Format (JSON):}
\begin{verbatim}
{
  "model_name": "{model_name}",
  "elite_or_not": true|false,
  "model_specific_patterns": [
    {
      "pattern_id": "M1_{model_name}",
      "pattern_name": "Model-specific pattern description",
      "pattern_type": "systematic_bias|confusion_pattern|...",
      "not_covered_by_common_patterns": true,
      "pattern_characterization": "How this model behaves differently",
      "examples": [
        "Context - MV: PER(\"...\"), {model_name}: MISC(\"...\")", ...
      ],
      "existing_instruction_assessment": {
        "covered_by_existing_model_specific": "true|false",
        "existing_instruction_reference": "instruction_id or null",
        "effectiveness_assessment": "qualitative note"
      }
    }
  ],
  "model_bias_summary": {
    "primary_systematic_biases": ["Bias 1", "Bias 2"],
    "model_strengths": ["Strength 1", "Strength 2"],
    "key_weaknesses": ["Weakness 1", "Weakness 2"],
    "deviation_from_mv_coalition": "narrative summary"
  }, ...
  "instruction_candidate_needs_max": "{max_model_specific_instructions}"
}
\end{verbatim}\\
\bottomrule
\end{tabular}
\caption{Prompt for instruction refinement in phase 2 (Single-Model Error Analysis).}
\label{tab:phase2_prompt}
\end{table*}

\begin{table*}[t!]
\centering
\rowcolors{1}{lightgray}{white}
\begin{tabular}{p{0.96\textwidth}}
\toprule
\textbf{Prompt Text} \\
\midrule
You are a strict, methodical NER annotation supervisor acting as an instruction generator and conflict resolver. Convert identified patterns into concrete instructions, resolve conflicts using the final task goal guidance, and run in either human-interactive or GPT-autonomous mode.

\textbf{Inputs}

- Phase 1 results: \texttt{{phase1\_results}}

- Phase 2 results: \texttt{{phase2\_results}}

- Final task goal: \texttt{{final\_task\_goal}}

- Decision mode: \texttt{{decision\_mode}} \;(\texttt{human\_interactive} | \texttt{gpt\_autonomous})

- Human input if interactive: \texttt{{human\_input}}

- Existing instructions: \texttt{{existing\_common\_instructions}}, \texttt{{existing\_model\_instructions}}

\textbf{Dynamic Parameters}

- \texttt{{max\_common\_instructions}}, \texttt{{max\_model\_specific\_instructions}}

\textbf{Output Format (JSON)}:
\begin{verbatim}
{
  "updated_final_goal": {
    "goal_updated": "true|false",
    "sections_updated": ["..."],
    "updated_final_goal_text": "..."
  },
  "decision_mode_used": "human_interactive|gpt_autonomous",
  "conflict_resolutions": [
    {
      "conflict_id": "C1",
      "conflicting_candidates": ["Candidate A", "Candidate B"],
      "resolution_rationale": "Why this choice was made", ...
    }
  ],
  "finalized_common_instructions": [
    {
      "instruction_id": "CI1",
      "instruction_text": "Concrete common instruction",
      "addresses_patterns": ["P1", "P2"],
      "examples": [ ... ],
      "priority": "high|medium|low",
      "instruction_type": "new|improved|replacement", ...
    }
  ],
  "finalized_common_instructions_max": "{max_common_instructions}",
  "finalized_model_instructions": {
    "{model_name}": [ ... ]
  },
  "finalized_model_instructions_max_per_model": 
    "{max_model_specific_instructions}",
  "instruction_generation_summary": ...  
}
\end{verbatim}\\
\bottomrule
\end{tabular}
\caption{Prompt for instruction refinement in phase 3 (Instruction Generation and Decision).}
\label{tab:phase3_prompt}
\end{table*}

\begin{table*}[t!]
\centering
\rowcolors{1}{lightgray}{white}
\begin{tabular}{p{0.96\textwidth}}
\toprule
\textbf{Prompt Text} \\
\midrule
You are a strict, methodical NER annotation supervisor acting as a guideline architect. Organize all instructions (existing + new) into a clear hierarchy, resolve remaining inconsistencies, and create the final guideline for the next iteration. Prioritize preservation of existing instructions and integrate new ones harmoniously.

\textbf{Inputs}
- Phase 3 results: \texttt{{phase3\_results}}

- Existing instructions: \texttt{{existing\_instructions}}

- Original NER scheme: \texttt{{current\_ner\_schema}}

- Updated final task goal: \texttt{{updated\_final\_goal}}

\textbf{Dynamic Parameters}
- \texttt{{preserve\_existing\_instructions}}, \texttt{{limit\_instruction\_changes}}, \texttt{{max\_change\_ratio}}

\textbf{Output Format (JSON)}:
\begin{verbatim}
{
  "instruction_integration_analysis": {
    "existing_instructions_retained": ...
    "preservation_score": "0.0-1.0"
  },
  "contradiction_resolutions": [
    {
      "contradiction_id": "CR1",
      "conflicting_instructions": ["Instruction A", "Instruction B"], ...
    }
  ],
  "hierarchical_common_instructions": [
    {
      "level": "1",
      "instruction_number": "1",
      "instruction_text": "Top-level principle", ...
      "sub_instructions": [
        {
          "level": "1.1",
          "instruction_number": "1.1",
          "instruction_text": "Sub-principle",
          "examples": [
            {"text": "Example", "correct_annotation": "Gold", "explanation": "Note"}
          ], ... }]}],
  "prioritized_model_instructions": {
    "{model_name}": [
      {
        "priority_rank": 1,
        "instruction_id": "MI1_{model_name}",
        "instruction_text": "Highest-priority instruction", ...
      }]},
  "final_guideline_summary": {
    "total_hierarchical_common_instructions": 0,
    "max_hierarchy_depth": 2,
    ...
  }
}
\end{verbatim}\\
\bottomrule
\end{tabular}
\caption{Prompt for instruction refinement in phase 4 (Hierarchical Guideline Organization.)}
\label{tab:phase4_prompt}
\end{table*}

\section{Methodology for Instruction Categorization}
\label{app:instruction_analysis}

To identify which instructions were introduced via inter-model disagreement and to assess their effectiveness across all 18 NER benchmarks, we categorized each ``Common Instruction'' generated by the supervisor model. This categorization helps explain how disagreement-guided refinement leads to performance gains. The categories are defined as follows:

\begin{itemize}
    \item \textbf{Span Boundary \& Composition}: Rules for entity extent, including modifiers and punctuation within a span.
    \item \textbf{Entityhood \& Referentiality}: Criteria for distinguishing entities from common nouns or generic mentions.
    \item \textbf{Type Disambiguation Logic}: Heuristics to resolve confusion between similar or overlapping entity types.
    \item \textbf{Global Strategy \& Purpose}: Instructions defining the task’s overarching goal and guiding philosophy.
    \item \textbf{Formatting \& Noise Handling}: Rules for handling symbols, tokenization artifacts, and orthographic noise.
    \item \textbf{Annotator Workflow \& Priority}: Guidance on decision-making sequences and rule precedence.
    \item \textbf{Others \& Specialized}: Niche domain-specific technical rules that do not fit into other categories.
\end{itemize}

\section{Supplementary Results}
\begin{figure*}[htbp]
    \centering
    \includegraphics[width=\linewidth]{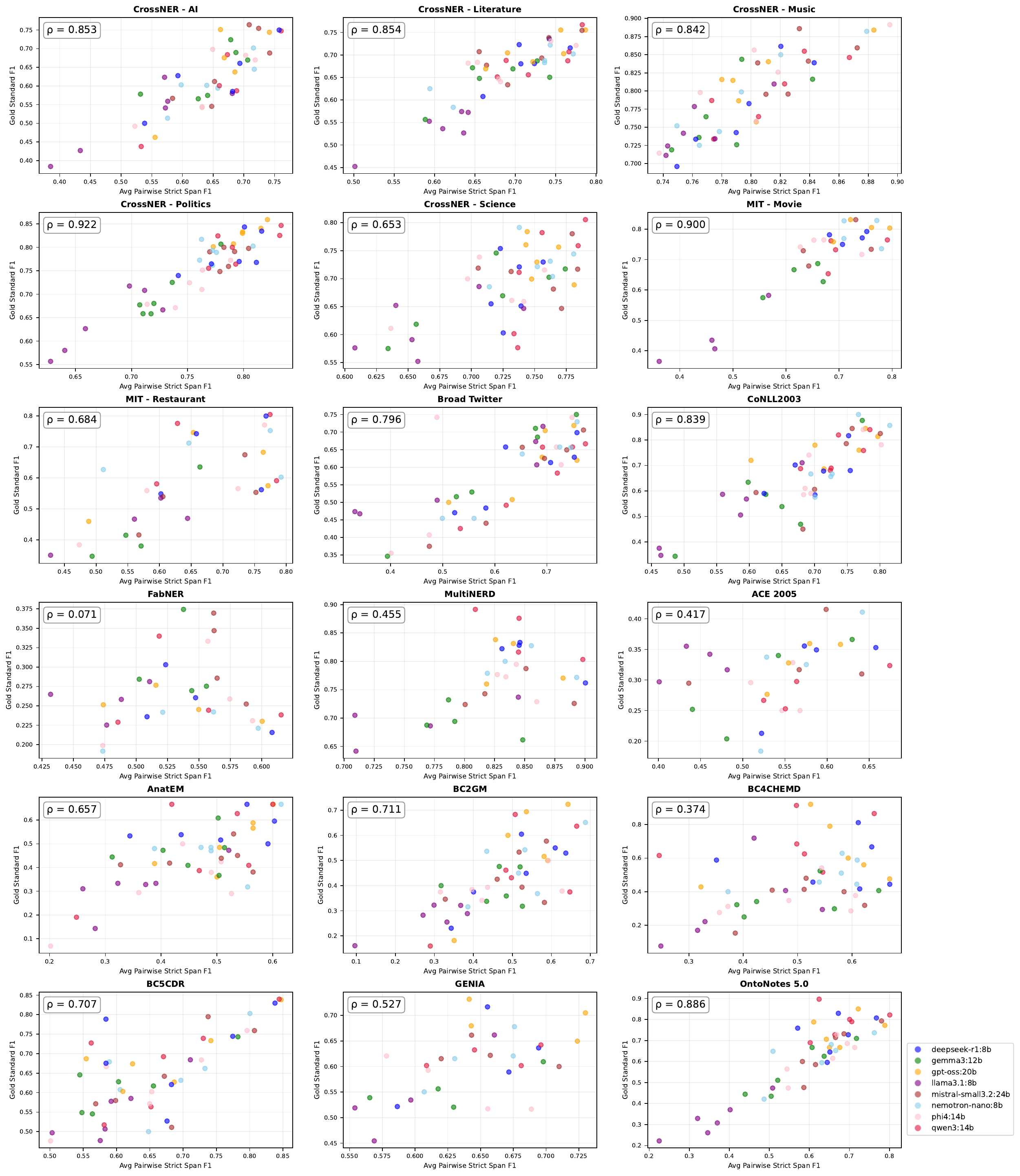}
    \caption{
        Correlation between pairwise agreement and NER performance across training iterations.Each subplot represents individual benchmarks, showing the Pearson correlation ($\rho$) between inter-annotator agreement (x-axis) and F1 performance on iteration document sets (y-axis). Strong positive correlations across diverse domains confirm that agreement statistics serve as a reliable, label-free indicator of NER performance during DiZiNER cycles.
    }
    \label{fig:agreement_correlation}
\end{figure*}

\begin{figure*}[htbp]
    \centering
    % Ensure the filename matches your provided path
    \includegraphics[width=0.95\linewidth]{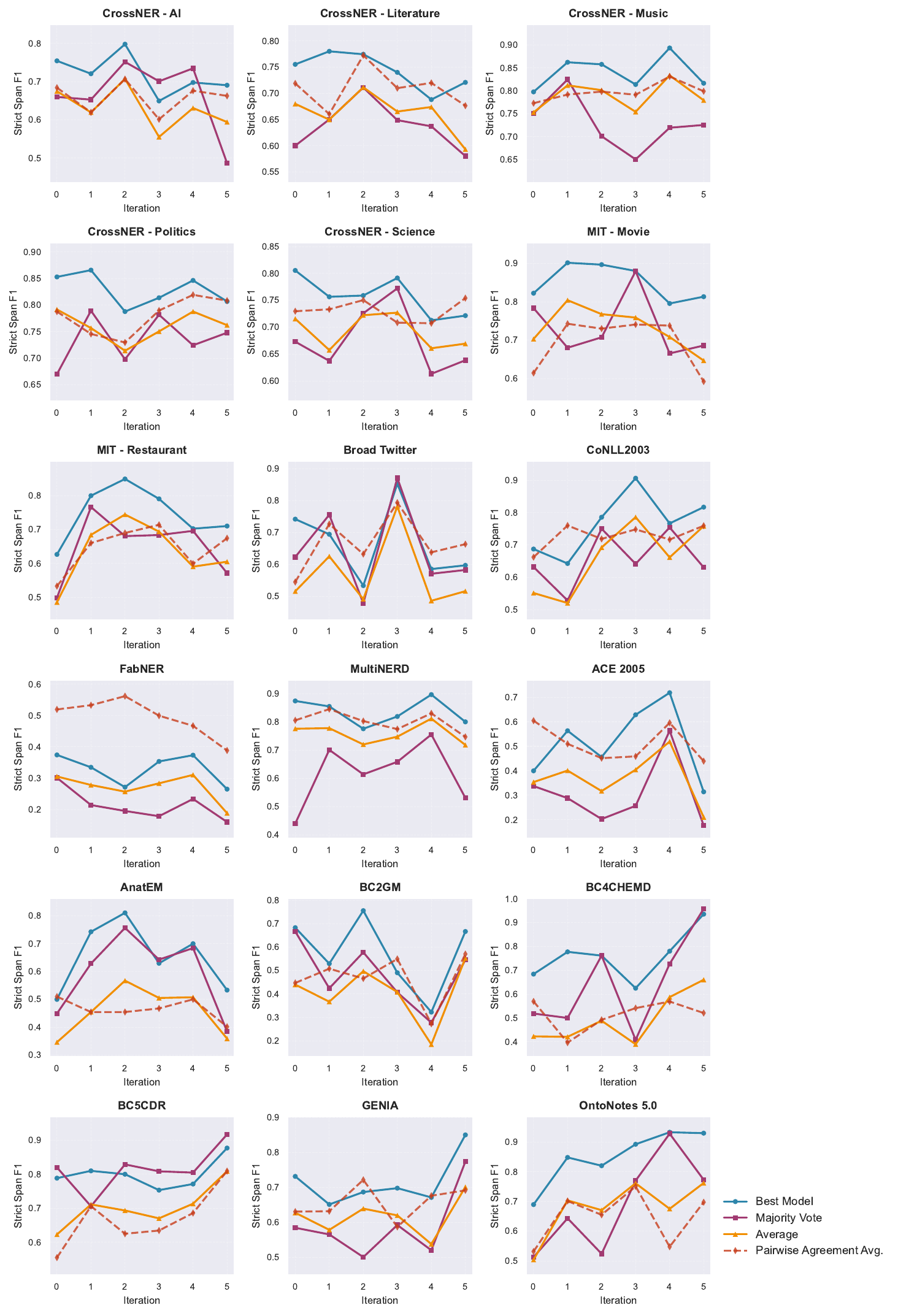}
    \caption{Evolution of NER performance and inter-annotator agreement across iterations for 18 benchmarks. Each plot displays the strict span f1 score measured on the iteration document sets for: (1) the Best Model (top-performing individual annotator), (2) the Majority Vote consensus, and (3) the Average performance of the eight heterogeneous LLM annotators. The dashed line (Pairwise Agreement Avg.) represents the mean inter-model agreement, demonstrating its role as a reliable, label-free proxy for performance gains during the DiZiNER cycles.}
    \label{fig:performance_evolution_full}
\end{figure*}

Figure~\ref{fig:agreement_correlation} visualizes the correlation between inter-annotator agreement and NER performance across refinement iterations. Figure~\ref{fig:performance_evolution_full} tracks the progression of inter-annotator agreement and NER performance on the iteration document sets across iterations for each individual benchmark.

Table~\ref{tab:ensemble_comparison_final} presents a comprehensive performance comparison against various ensemble methods, while Table~\ref{tab:diverse_supervisors} details the results across different supervisor models. The impact of the final task goal is examined in Table~\ref{tab:skipping_final_goal}.

Table~\ref{tab:abl_dropping_worst} shows the comparison with and without the removal of the least consistent annotator. Table~\ref{tab:abl_group_size} summarizes the results under different document set sizes per iteration. Furthermore, Table~\ref{tab:abl_llm_families} investigates performance across different annotator families, and Table~\ref{tab:performance_model_comparison} analyzes the effects of varying the number of annotator models. Table~\ref{tab:results_supervised} provides a comparison between zero-shot and supervised evaluation settings.

Finally, Table~\ref{tab:instruction_category_stats} provides a categorical distribution of the refined instructions, and Table~\ref{tab:qualitative_ner_results} showcases qualitative NER results across diverse benchmarks.

\begin{table*}[htbp]
\centering
\begin{tabular}{lccccccc}
\toprule
\textbf{Benchmark} & \textbf{DiZiNER} & \textbf{MV} & \textbf{DS} & \textbf{GLAD} & \textbf{MACE} & \textbf{GPT-5 mini} & \textbf{Prior Best ZS} \\
\midrule
\textbf{AI}           & 71.1 [71.1, 71.1] & \textbf{73.0} & 69.3 & \underline{72.3} & 71.9 & 64.3 & 68.2 \\
\textbf{Literature}   & \textbf{72.7 [72.0, 73.8]} & 69.3 & 66.8 & 69.4 & 69.1 & 67.6 & \underline{71.6} \\
\textbf{Music}        & 80.6 [79.2, 82.9] & 83.1 & 80.7 & \textbf{83.5} & \underline{83.3} & 73.3 & 82.4 \\
\textbf{Politics}     & \textbf{79.4 [77.6, 80.9]} & \underline{79.1} & 77.0 & 79.0 & 78.8 & 72.8 & 78.2 \\
\textbf{Science}      & \underline{74.8 [74.1, 75.4]} & 72.6 & 72.4 & 73.8 & 73.1 & 68.4 & \textbf{79.4} \\
\textbf{Movie}        & \textbf{76.2 [74.4, 78.5]} & \underline{74.2} & 71.1 & 73.8 & \underline{74.2} & 73.3 & 70.9 \\
\textbf{Restaurant}   & 67.3 [66.9, 68.1] & 66.2 & 62.6 & \underline{68.0} & 67.7 & 58.5 & \textbf{69.3} \\
\textbf{BroadTwitter} & \textbf{76.9 [75.5, 78.3]} & 67.8 & 55.9 & 62.3 & 61.5 & 59.2 & \underline{67.9} \\
\textbf{ACE05}        & 45.0 [44.1, 46.2] & 22.1 & 24.9 & 22.6 & 23.5 & \textbf{54.0} & \underline{51.2} \\
\textbf{CoNLL2003}    & 86.9 [85.8, 88.6] & \textbf{93.2} & 86.5 & \underline{92.9} & \underline{92.9} & 81.8 & 81.3 \\
\textbf{MultiNERD}    & 80.6 [79.0, 83.7] & \underline{81.5} & 68.3 & \textbf{81.8} & \underline{81.5} & 74.0 & 77.5 \\
\textbf{OntoNotes}    & \underline{62.5 [61.5, 63.5]} & 54.4 & 46.4 & 58.1 & 58.6 & \textbf{63.8} & 32.2 \\
\textbf{FabNER}       & 29.5 [28.9, 30.5] & 32.6 & \underline{33.4} & 33.2 & \textbf{33.7} & 29.5 & 26.3 \\
\textbf{AnatEM}       & 59.1 [56.1, 60.7] & 60.4 & 42.1 & \underline{62.8} & \textbf{63.8} & 59.2 & 33.3 \\
\textbf{BC2GM}        & \underline{71.0 [67.0, 73.0]} & 62.3 & 57.0 & 63.9 & 63.8 & \textbf{73.0} & 47.9 \\
\textbf{BC4CHEMD}     & \textbf{79.5 [78.4, 81.7]} & 71.6 & 54.9 & 70.0 & \underline{72.7} & 63.7 & 47.9 \\
\textbf{BC5CDR}       & \textbf{78.9 [77.0, 81.3]} & 77.1 & 65.9 & 77.5 & \underline{77.6} & 62.8 & 68.0 \\
\textbf{GENIA}        & \textbf{60.1 [59.6, 60.9]} & 58.0 & 54.2 & \underline{58.9} & 58.4 & 56.5 & 55.5 \\
\midrule
\textbf{Average}      & \textbf{69.6 [68.2, 71.1]} & 66.6 & 60.5 & 66.9 & \underline{67.0} & 64.2 & 61.6 \\
\bottomrule
\end{tabular}
\caption{Performance comparison across 18 NER benchmarks including DiZiNER, various ensemble methods, GPT-5 mini baseline, and previous zero-shot SOTA. The ensemble methods include Majority Voting (MV), Dawid-Skene (DS), Generative model of Labels, Abilities, and Difficulties (GLAD), and Multi-Annotator Competence Estimation (MACE). For DiZiNER ZS, values are reported as average F1 [min, max]. Best and second-best results for each individual benchmark are highlighted in \textbf{bold} and \underline{underlined}, respectively.}
\label{tab:ensemble_comparison_final}
\end{table*}

\begin{table*}[htbp]
\footnotesize

\centering
\begin{adjustbox}{width=0.6\linewidth}
\begin{tabular}{llc}
\toprule
\textbf{Benchmark} & \textbf{Supervisor} & \textbf{F1 [min, max]} \\
\midrule
\multirow{7}{*}{\textbf{AI}} 
 & \textit{gpt-5-mini-2025-08-07} & \textbf{71.1 [71.1, 71.1]} \\
 & \textit{gpt-oss-120b} & 65.1 [61.1, 67.3] \\
 & \textit{qwen-2.5-72b-instruct} & 65.6 [56.5, 70.5] \\
 & \textit{llama-3.3-70b-instruct} & 66.0 [62.5, 69.3] \\
 \cmidrule{2-3}
 & Prior Best ZS & \underline{68.2} \\
 & GPT-5 mini baseline & 64.3 \\
\midrule
\multirow{7}{*}{\textbf{Literature}} 
 %& \textit{gpt-5-2025-08-07} & 70.5 [69.1, 72.0] \\ % Removed
 & \textit{gpt-5-mini-2025-08-07} & \textbf{72.7 [72.0, 73.8]} \\
 & \textit{gpt-oss-120b} & 69.2 [65.2, 71.4] \\
 & \textit{qwen-2.5-72b-instruct} & 71.0 [69.2, 72.0] \\
 & \textit{llama-3.3-70b-instruct} & 69.0 [66.2, 71.7] \\
 \cmidrule{2-3}
 & Prior Best ZS & \underline{71.6} \\
 & GPT-5 mini baseline & 67.6 \\
\bottomrule
\end{tabular}
\end{adjustbox}
\vspace{4pt}
\caption{Performance comparison of DiZiNER across various supervisor models, alongside GPT-5 mini baselines and prior best results on CrossNER-AI and Literature. For DiZiNER configurations, values are reported as average F1 [min, max]. Best and second-best results for each individual benchmark are highlighted in \textbf{bold} and \underline{underlined}, respectively.}
\label{tab:diverse_supervisors}
\end{table*}

\begin{table*}[htbp]
\centering
\begin{tabular}{lcc}
\toprule
\textbf{Benchmark} & \textbf{Base} & \textbf{+sfg} \\
\midrule
\textbf{AI}          & \textbf{71.1 [71.1, 71.1]} & 65.0 [62.3, 68.3] \\
\textbf{Literature} & \textbf{72.7 [72.0, 73.8]} & 65.1 [62.8, 66.8] \\
\textbf{Music}      & \textbf{80.6 [79.2, 82.9]} & 75.8 [71.0, 78.9] \\
\textbf{Politics}   & \textbf{79.4 [77.6, 80.9]} & 72.4 [67.2, 76.4] \\
\textbf{Science}    & \textbf{74.8 [74.1, 75.4]} & 70.8 [68.1, 73.4] \\
\textbf{CoNLL2003}  & \textbf{86.9 [85.8, 88.6]} & 82.1 [81.5, 82.8] \\
\midrule
\textbf{Average} & \textbf{77.6 [76.6, 78.8]} & 71.9 [68.8, 74.4] \\
\bottomrule
\end{tabular}
\caption{Best NER performance with and without skipping the final task goal (+sfg). Values are reported as average
F1 [min, max]. Best for each individual benchmark is highlighted in \textbf{bold}.}
\label{tab:skipping_final_goal}
\end{table*}

\begin{table*}[htbp]
\centering
\begin{tabular}{lccc}
\toprule
\textbf{Benchmark} & \textbf{Base} & \textbf{+dwa} & \textbf{\boldmath$\Delta_{\textit{Base - dwa}}$} \\
\midrule
\textbf{AI}           & \textbf{71.1 [71.1, 71.1]} & 69.9 [69.4, 70.4] & +0.0 \\
\textbf{Literature}   & \textbf{72.7 [72.0, 73.8]} & 72.6 [71.9, 73.8] & +0.1 \\
\textbf{Music}        & \textbf{80.6 [79.2, 82.9]} & 80.5 [80.0, 81.3] & +0.1 \\
\textbf{Politics}     & \textbf{79.4 [77.6, 80.9]} & 77.3 [76.6, 77.7] & +2.1 \\
\textbf{Science}      & \textbf{74.8 [74.1, 75.4]} & 73.9 [73.5, 74.7] & +0.9 \\
\textbf{Movie}        & \textbf{76.2 [74.4, 78.5]} & 70.2 [65.5, 73.7] & +6.0 \\
\textbf{Restaurant}   & 67.3 [66.9, 68.1] & \textbf{68.3 [67.8, 68.6]} & -1.0 \\
\textbf{BroadTwitter} & \textbf{76.9 [75.5, 78.3]} & 66.5 [60.7, 69.5] & +10.4 \\
\textbf{ACE05}        & \textbf{45.0 [44.1, 46.2]} & 40.3 [35.4, 46.2] & +4.7 \\
\textbf{CoNLL2003}    & \textbf{86.9 [85.8, 88.6]} & 86.7 [85.8, 88.6] & +0.2 \\
\textbf{MultiNERD}    & \textbf{80.6 [79.0, 83.7]} & 79.0 [78.4, 79.6] & +1.6 \\
\textbf{OntoNotes}    & 62.5 [61.5, 63.5] & \textbf{62.8 [62.1, 63.2]} & -0.3 \\
\textbf{FabNER}       & \textbf{29.5 [28.9, 30.5]} & 27.9 [25.9, 29.1] & +1.6 \\
\textbf{AnatEM}       & \textbf{59.1 [56.1, 60.7]} & 55.1 [54.6, 55.6] & +4.0 \\
\textbf{BC2GM}        & \textbf{71.0 [67.0, 73.0]} & 65.5 [60.1, 69.6] & +5.5 \\
\textbf{BC4CHEMD}     & \textbf{79.5 [78.4, 81.7]} & 79.1 [75.7, 81.7] & +0.4 \\
\textbf{BC5CDR}       & 78.9 [77.0, 81.3] & \textbf{80.9 [80.2, 81.3]} & -2.0 \\
\textbf{GENIA}        & \textbf{60.1 [59.6, 60.9]} & 57.8 [56.9, 59.5] & +2.3 \\
\midrule
\textbf{Average}      & \textbf{69.6 [68.2, 71.1]} & 67.5 [65.6, 69.1] & \textbf{+2.1} \\
\bottomrule
\end{tabular}
\caption{Zero-shot NER performance of DiZiNER with and without dropping the worst annotator (dwa). The worst model is defined as the model showing the highest average disagreement with all others. Values are reported
as average F1 [min, max]. Best for each individual benchmark is highlighted in \textbf{bold}.}
\label{tab:abl_dropping_worst}
\end{table*}

\begin{table*}[htbp]
\centering
\small
\begin{tabular}{lcccc}
\toprule
\multirow{2}{*}{\textbf{Benchmark}} & \multicolumn{4}{c}{\textbf{Iteration Document Set Size}} \\
\cmidrule{2-5}
 & \textbf{15} & \textbf{25} & \textbf{50} & \textbf{100} \\
\midrule
\textbf{AI}         & \underline{70.5 [67.9, 72.4]} & \textbf{71.1 [71.1, 71.1]} & 67.2 [65.8, 68.0] & 70.3 [69.4, 71.4] \\
\textbf{Literature} & \underline{70.6 [68.9, 72.5]} & \textbf{72.7 [72.0, 73.8]} & 69.8 [68.7, 71.8] & 69.0 [67.5, 71.9] \\
\midrule
\textbf{Average}    & \underline{70.6 [68.4, 72.5]} & \textbf{71.9 [71.6, 72.5]} & 68.5 [67.3, 69.9] & 69.7 [68.5, 71.7] \\
\bottomrule
\end{tabular}
\caption{Zero-shot NER performance across different iteration document set sizes. Values are reported as average F1 [min, max]. Best and second-best results for each individual benchmark are highlighted in \textbf{bold} and \underline{underlined}, respectively.}
\label{tab:abl_group_size}
\end{table*}

\begin{table*}[htbp]
\centering
\small
\setlength{\tabcolsep}{12pt}
\renewcommand{\arraystretch}{1.2}
\begin{adjustbox}{max width=\textwidth}
\begin{tabular}{lccc}
\toprule
\textbf{Benchmark} & \textbf{Base} & \textbf{Qwen} & \textbf{Llama} \\
\midrule
\textbf{AI}         & \textbf{71.1 [71.1, 71.1]} & 68.1 [66.1, 71.1] & 69.4 [68.8, 70.2] \\
\textbf{Literature} & \textbf{72.7 [72.0, 73.8]} & 68.2 [65.7, 70.6] & 71.0 [70.9, 71.1] \\
\midrule
\textbf{Average}    & \textbf{71.9 [71.6, 72.5]} & 68.2 [65.9, 70.9] & 70.2 [69.9, 70.7] \\
\bottomrule
\end{tabular}
\end{adjustbox}
\caption{Best zero-shot NER performance across different annotator families. The Base configuration utilizes eight heterogeneous models (all $\leq$ 24B parameters). In contrast, the Qwen and Llama configurations consist of eight models from their respective single families, including significantly larger models such as Llama 3.3-70B and Qwen 2.5-Coder-32B to meet the count requirement. Values are reported as average F1 [min, max]. Best for each individual benchmark is highlighted in \textbf{bold}.}
\label{tab:abl_llm_families}
\end{table*}

\begin{table*}[htbp]
\centering
\small
\begin{tabular}{lcccc}
\toprule
\multirow{2}{*}{\textbf{Benchmark}} & \multicolumn{4}{c}{\textbf{Number of Annotator Models}} \\
\cmidrule{2-5}
 & \textbf{4} & \textbf{8} & \textbf{12} & \textbf{16} \\
\midrule
\textbf{AI}         & \underline{68.7 [67.5, 69.6]} & \textbf{69.9 [68.7, 71.1]} & 68.4 [66.9, 69.6] & 67.4 [66.9, 68.0] \\
\textbf{Literature} & \underline{72.3 [72.0, 72.7]} & \textbf{72.7 [72.0, 73.8]} & 70.4 [66.4, 73.0] & 70.2 [68.7, 71.0] \\
\textbf{Music}      & 76.6 [75.3, 78.8] & \underline{80.6 [79.2, 82.9]} & \textbf{80.8 [79.2, 81.6]} & 80.2 [79.6, 80.6] \\
\textbf{Politics}   & 76.5 [76.0, 77.4] & \textbf{79.4 [77.6, 80.9]} & \underline{77.7 [75.7, 79.9]} & 77.3 [76.0, 78.6] \\
\textbf{Science}    & 71.5 [64.0, 75.8] & \textbf{74.8 [74.1, 75.4]} & 72.1 [69.5, 73.5] & \underline{74.6 [73.4, 75.8]} \\
\midrule
\textbf{Average}    & 73.1 [71.0, 74.9] & \textbf{75.5 [74.3, 76.8]} & \underline{73.9 [71.5, 75.5]} & \underline{73.9 [72.9, 74.8]} \\
\bottomrule
\end{tabular}
\caption{Zero-shot NER performance across different numbers of annotator models. Best and second-best results for each individual benchmark are highlighted in \textbf{bold} and \underline{underlined}, respectively.}
\label{tab:performance_model_comparison}
\end{table*}

\begin{table*}[htbp]
\centering
\begin{tabular}{lccc}
\toprule
\textbf{Benchmark} & \textbf{Zero-shot} & \textbf{Supervised} & \textbf{\boldmath$\Delta_{\textit{Sup. - ZS}}$} \\
\midrule
\textbf{AI}          & 69.9 [68.7, 71.1] & \textbf{70.1 [68.8, 71.1]} & +0.2 \\
\textbf{Literature}  & \textbf{72.7 [72.0, 73.8]} & 71.1 [67.0, 73.5] & -1.6 \\
\textbf{Music}       & \textbf{80.6 [79.2, 82.9]} & 79.2 [78.6, 79.9] & -1.4 \\
\textbf{Politics}    & \textbf{79.4 [77.6, 80.9]} & 76.4 [76.2, 76.6] & -3.0 \\
\textbf{Science}     & \textbf{74.8 [74.1, 75.4]} & 71.9 [71.6, 72.3] & -2.9 \\
\textbf{Movie}       & 76.2 [74.4, 78.5] & \textbf{77.1 [75.0, 78.7]} & +0.9 \\
\textbf{Restaurant}  & 67.3 [66.9, 68.1] & \textbf{69.3 [67.0, 72.7]} & +2.0 \\
\textbf{BroadTwitter}& \textbf{76.9 [75.5, 78.3]} & 74.3 [73.8, 75.2] & -2.6 \\
\textbf{ACE05}       & 45.0 [44.1, 46.2] & \textbf{55.5 [55.0, 55.8]} & +10.5 \\
\textbf{CoNLL2003}   & \textbf{86.9 [85.8, 88.6]} & 86.1 [84.1, 87.9] & -0.8 \\
\textbf{MultiNERD}   & 80.6 [79.0, 83.7] & \textbf{82.2 [81.5, 82.8]} & +1.6 \\
\textbf{OntoNotes}   & 62.5 [61.5, 63.5] & \textbf{68.1 [66.4, 69.0]} & +5.6 \\
\textbf{FabNER}      & \textbf{29.5 [28.9, 30.5]} & 28.4 [28.0, 29.0] & -1.1 \\
\textbf{AnatEM}      & 59.1 [56.1, 60.7] & \textbf{59.4 [57.1, 60.9]} & +0.3 \\
\textbf{BC2GM}       & \textbf{71.0 [67.0, 73.0]} & 68.4 [67.2, 69.4] & -2.6 \\
\textbf{BC4CHEMD}    & 79.5 [78.4, 81.7] & \textbf{80.1 [79.5, 80.9]} & +0.6 \\
\textbf{BC5CDR}      & 78.9 [77.0, 81.3] & \textbf{80.8 [79.9, 81.8]} & +1.9 \\
\textbf{GENIA}       & \textbf{60.1 [59.6, 60.9]} & 57.5 [53.8, 61.1] & -2.6 \\
\midrule
\textbf{Average} & 69.5 [67.0, 71.1] & \textbf{69.8 [67.9, 71.6]} & \textbf{+0.3} \\
\bottomrule
\end{tabular}
\caption{Comparison of DiZiNER zero-shot and supervised performance across benchmarks. Best for each individual benchmark is highlighted in \textbf{bold}.}
\label{tab:results_supervised}
\end{table*}

\begin{table*}[htbp]
\centering
\begin{tabular}{lccc}
\toprule
\textbf{Category} & \textbf{High-perf. Conf.} & \textbf{Low-perf. Conf.} & \textbf{Difference} \\
\midrule
Span Boundary \& Composition & 2.9 (29.4\%) & 2.6 (27.6\%) & +0.3 (+1.8\%) \\
Entityhood \& Referentiality  & 2.3 (23.4\%) & 1.8 (18.7\%) & +0.5 (+4.7\%) \\
Type Disambiguation Logic    & 2.1 (20.7\%) & 2.8 (29.0\%) & -0.7 (-8.3\%) \\
Global Strategy \& Purpose    & 1.3 (12.7\%) & 1.0 (9.9\%)  & +0.3 (+2.8\%) \\
Formatting \& Noise Handling  & 0.6 (6.2\%)  & 0.6 (6.0\%)  & +0.0 (+0.2\%) \\
Annotator Workflow \& Priority & 0.5 (4.7\%)  & 0.6 (6.4\%)  & -0.1 (-1.7\%) \\
Others \& Specialized         & 0.4 (3.8\%)  & 0.3 (3.2\%)  & +0.1 (+0.6\%) \\
\midrule
\textbf{Total}                               & \textbf{10.0} & \textbf{9.6} & \textbf{+0.4} \\
\bottomrule
\end{tabular}
\caption{Comparison of category distributions of refined common instruction between high-performing and low-performing iteration--model configurations.}
\label{tab:instruction_category_stats}
\end{table*}

\begin{table*}[htbp]
\centering
\small
\setlength{\tabcolsep}{5pt}
\renewcommand{\arraystretch}{1.5}
\resizebox{\textwidth}{!}{
\begin{tabular}{p{4.5cm}p{2.5cm}p{2cm}p{2cm}p{4cm}}
\toprule
\textbf{Input Text [Benchmark]} & \textbf{Gold Entity} & \textbf{DiZiNER} & \textbf{Iteration 0} & \textbf{Remarks} \\
\midrule
@scotthornsby10 to be clear, it's only for people, not brands. [Broad Twitter] & scotthornsby10 (PER) & scotthornsby10 & @scotthornsby10 & Successfully removed @ symbols to align with standard person mention spans. \\
Ex vivo, estradiol exposure increased the IL-8 secretion of normal whole breast tissue in culture. [AnatEM] & breast tissue (ANAT) & breast tissue & normal whole breast tissue & Excluded descriptive modifiers to isolate the core anatomical entity within the span. \\
Cambridge 22 13 3... [CoNLL2003] & Cambridge (ORG) & Cambridge & Cambridge (LOC) & Used league table context to correctly classify the city name as an organization. \\
G-CSF (10 microg/kg) was started on day + 1 and all patients engrafted... [BC2GM] & G-CSF (GENE) & G-CSF & None & Detected a technical gene abbreviation. \\
are there any places left that allow smoking in a restaurant [MIT Restaurant] & allow smoking (AMEN) & allow smoking & None & Captured a long-form descriptive functional entity. \\
so are you going to get an article in Nature or something? [OntoNotes] & Nature (ORG) & Nature & None & Identified a specific domain publication name previously missed in the initial result. \\
Assessment of the abuse liability of ABT-288, a novel histamine H3 receptor antagonist. [BC4CHEMD] & ABT-288 (CHEM), histamine (CHEM) & ABT-288, histamine & ABT-288 & Identified missing chemical mentions. \\
In the busy Fucheng district, you find the Taiwanese bars, covered from door to rooftop with flashing lights [OntoNotes] & Fucheng (GPE) & Fucheng & Fucheng district & Removed generic district markers to isolate the specific geographical name as a GPE. \\
A single grid can be analysed for both content (eyeball inspection)... [AI] & eyeball inspection (TASK) & eyeball inspection & None & Captured specific task entities. \\
A confusion matrix or matching matrix is often used as a tool to validate the accuracy of k-NN classification. [AI] & accuracy (METR), k-NN classification (ALG) & accuracy, k-NN classification & k-NN classification & Improved recall for evaluation metrics within technical algorithmic descriptions. \\
\bottomrule
\end{tabular}
}
\caption{Qualitative NER results of DiZiNER compared with the results of Iteration 0 using the same annotator model and input text.}
\label{tab:qualitative_ner_results}
\end{table*}

\end{document}